\documentclass[conference]{IEEEtran}
\IEEEoverridecommandlockouts
\usepackage{cite}
\usepackage{amsmath,amssymb,amsfonts}
\usepackage{graphicx}
\usepackage{textcomp}
\usepackage{xcolor}

\usepackage{balance}
\usepackage{url}
\usepackage{algorithm}
\usepackage{algorithmic}
\usepackage{caption}
\usepackage{subcaption}
\usepackage{comment}
\usepackage{setspace}
\usepackage{enumerate}
\usepackage{amsthm}
\usepackage{xspace}
\usepackage{graphicx}
\usepackage[export]{adjustbox}
\usepackage{subcaption}

\newcommand{\model}{\textbf{M-Attack}\xspace}  
\newcommand{\distance}{\textit{Mix-Mahalanobis Distance}\xspace}

\def\BibTeX{{\rm B\kern-.05em{\sc i\kern-.025em b}\kern-.08em
    T\kern-.1667em\lower.7ex\hbox{E}\kern-.125emX}}
\begin{document}

\title{Towards Generating Adversarial Examples on Mixed-type Data}

\author{\IEEEauthorblockN{Han Xu}
\IEEEauthorblockA{\textit{Michigan State University, East Lansing, Michigan, USA} \\
xuhan1@msu.edu}\\
\IEEEauthorblockN{Zhimeng Jiang}
\IEEEauthorblockA{\textit{Texas A\&M University, College Station, Texas, USA} \\
zhimengj@tamu.edu}
\\
\IEEEauthorblockN{Menghai Pan, Huiyuan Chen, Xiaoting Li, Mahashweta Das, Hao Yang}
\IEEEauthorblockA{\textit{VISA Research, Palo Alto, California, USA} \\
}
}

\maketitle

\begin{abstract}
The existence of adversarial attacks (or adversarial examples) brings huge concern about the machine learning (ML) model's safety issues. For many safety-critical ML tasks, such as financial forecasting, fraudulent detection, and anomaly detection, the data samples are usually mixed-type, which contain plenty of numerical and categorical features at the same time. However, how to generate adversarial examples with mixed-type data is still seldom studied. In this paper, we propose a novel attack algorithm \model, which can effectively generate adversarial examples in mixed-type data. Based on \model, attackers can attempt to mislead the targeted classification model's prediction, by only slightly perturbing both the numerical and categorical features in the given data samples. More importantly, by adding designed regularizations, our generated adversarial examples can evade potential detection models, which makes the attack indeed insidious. Through extensive empirical studies, we validate the effectiveness and efficiency of our attack method and evaluate the robustness of existing classification models against our proposed attack. The experimental results highlight the feasibility of generating adversarial examples toward machine learning models in real-world applications.

\end{abstract}

\section{Introduction}

The existence of adversarial attacks (or adversarial examples)~\cite{madry2017towards, goodfellow2014explaining, tramer2019adversarial, wei2019transferable} has brought huge concerns when applying machine learning (ML) models to safety-critical tasks. 
In many ML applications in Web Service or mobile applications, the data inputs are often mixed type, which contains both numerical and categorical features simultaneously. For example, for an online financial institute, ML models are trained to evaluate whether a loan applicant has the ability to repay his/her loan.  In this scenario, the data has numerical features, e.g., the applicant's age, account balance, and annual income, as well as categorical features including his/her educational background, occupation type, and marital status. Similarly, in recommender systems for online shopping, the data also contains both numerical and categorical information, such as the price and categories of the recommended items. However, how to conduct adversarial attacks for mixed-type data still lacks full exploration. Therefore, in this paper, we focus on the problem of how to generate adversarial examples for mixed-type data. Specifically, we aim to solve the problem: \textit{given a well-trained ML model, how can we perturb a data sample with an unnoticeable distortion, such that this ML model's prediction is misled to give a wrong prediction?} The studied problem is crucial in practice. Recall the above financial institute example: if an unqualified applicant provides a fake profile that contains a few fraudulent records so that he/she can fool the trained ML model to get the approval of the loan, it will cause a huge cost to this financial institute.

To achieve this attacking goal, we face tremendous challenges. 
First, the attacker is expected to modify as few features of the original (clean) data as possible to keep the perturbation behavior unnoticeable and insidious. This requires the perturbation added on the original sample to \textbf{be sparse in the input data space}. Notably, there are existing methods for sparse adversarial attacks in either numerical or categorical data domains, separately. In the numerical domain, Projected Gradient Descent methods (PGD) are adopted to guide the search of adversarial examples and project the perturbation into a continuous $l_1$-norm bounded space to stress the sparsity of perturbation~\cite{madry2017towards, maini2020adversarial}. In the categorical domain, search-based methods~\cite{yang2020greedy, bao2021towards} are employed to iteratively find the top-$K$ categorical features, which have the largest influence to model prediction, and then search optimal perturbation for these $K$ features. However, our task involves both the numerical and categorical features and there are still no confirmative methods to generate the optimal sparse adversarial attack over the targeted searching space, which is a Cartesian product of a discrete space (for categorical features) and continuous space (for numerical features). Moreover, our experimental results in Section~\ref{sec:exp} suggest that simply combining existing strategies usually provide sub-optimal solutions. For example, for an algorithm which first applies the search-based methods~\cite{yang2020greedy, bao2021towards} to perturb categorical features, and then applies $l_1$-PGD method~\cite{madry2017towards, maini2020adversarial} to find numerical perturbations, it cannot successfully find strongest adversarial examples (or leading to the maximal loss value of the targeted classifier). This fact highlights the necessity to devise new adversarial attack algorithms exclusively designed for mixed-type data.

Second, the attacker should also \textbf{keep the generated adversarial examples to be seemingly benign}. In other words, adversarial examples should be close to the distribution of original clean data samples. It is hard to achieve this goal since features in mixed-type data are usually highly correlated. For example, in Home Credit Default Risk Dataset~\cite{ozdemir2004empirical} where the task is to predict a person's qualification for a loan, the feature ``age'' of an applicant is always strongly related to other numerical features (such as ``number of children'') and categorical features (such as ``family status''). In this dataset, if an attacker perturbs the feature ``family status'' from ``child'' to ``parent'' for an 18-year-old loan applicant, the perturbed sample obviously deviates from the true distribution of clean data samples (because there are rarely 18-year-old parents in reality). As a result, the generated adversarial example can be easily detected as ``abnormal'' samples, by an defender that applies Out-of-Distribution (OOD) detection (or Anomaly Detection) methods~\cite{liu2020energy, botev2010kernel, nalisnick2018deep}, or detected by human experts who can judge the authentication of data samples based on their domain knowledge. Thus, the attacker should generate adversarial examples, which do not significantly violate the correlation between any pair of numerical features, as well as any pair of categorical features, or the pair of categorical and numerical features.

In this paper, to tackle the aforementioned challenges, we proposed \model, which is a novel attacking framework for mixed-type data. In detail, we first transform the searching space of mixed-type adversarial examples into \textit{a unified continuous space} (see Figure~\ref{fig:frame}). To achieve this goal, we convert the problem of finding adversarial categorical perturbations into the problem to find the probabilistic (categorical) distribution which the adversarial categorical features are sampled from. Therefore, we are facilitated to find sparse adversarial examples in this unified continuous space, by simultaneously updating the numerical \& categorical perturbation via gradient-based methods. Furthermore, to generate in-distribution adversarial examples, we propose a \distance to measure and regularize the distance of an (adversarial) example to the distribution of the clean mixed-type data samples. Through extensive experiments, we verify that: (1) \model can achieve better attacking performance compared to representative baselines, which are the combination of existing numerical \& categorical attack methods; (2) \model can achieve better (or similar) efficiency compared to these baseline methods; and (3) \distance can help the generated adversarial examples be close to the true distribution of clean data. Our contribution can be summarized as follows:
\begin{itemize}
    \item We propose an efficient and effective algorithm, \textbf{M-Attack}, to generate adversarial examples for mixed-type data. 
    \item We propose a novel measurement, Mixed Mahalanobis Distance, which helps the generated adversarial examples be close to the true distribution. 
    \item We conduct extensive experiments to validate the feasibility of \textbf{M-Attack}, and demonstrate the vulnerability of popular classification models against \textbf{M-Attack}. 
\end{itemize}
\section{Related Works}

There has been a rise in the importance of the adversarial robustness of machine learning models in recent years. Most of the existing works, including evasion attack, and poison attack, focus on continuous input space, especially in the image domain~\cite{shafahi2018poison, madry2017towards, ilyas2019adversarial}. In the image domain, because the data space (which is the space of pixel values) is continuous, people use gradient-based methods~\cite{madry2017towards, maini2020adversarial} to find adversarial examples. Based on the gradient attack methods, various defense methods, such as adversarial training,~\cite{madry2017towards, zhang2019theoretically} are proposed to improve the model robustness.
Meanwhile, adversarial attacks focusing on discrete input data, like text data, which have categorical features, are also starting to catch the attention of researchers. In the natural language processing domain, the work \cite{kuleshov2018adversarial} discusses the potential to attack text classification models such as sentiment analysis, by replacing words with their synonyms. The study \cite{ebrahimi2017hotflip} proposes to modify the text token based on the gradient of input one-hot vectors. The method \cite{gao2018black} develops a scoring function to select the most effective attack and a simple character-level transformation to replace projected gradient or multiple linguistic-driven steps on text data. In the domain of graph data learning, there are methods~\cite{jin2021adversarial} that greedily search the perturbations to manipulate the graph structure to mislead the targeted models. In a conclusion, when the data space is discrete, these methods share a similar core idea by applying searched-based methods to find adversarial perturbations. Although there are well-established methods for either numerical domain or categorical domain separately, there is still a lack of studies in mixed-type data. However, mixed-type data are widely existing in various machine learning tasks in the physical world. For example, for the fraudulent transaction detection systems~\cite{bolton2002statistical, ghosh1994credit} of credit card institutes, the transaction records from cardholders may include features such as transaction amount (numerical) and the type of the purchased product (categorical). Similarly, for ML models applying in AI healthcare~\cite{panch2019inconvenient, reddy2020governance}, i.e., in epidemiological studies~\cite{rothman2008modern, rothman2012epidemiology}, the data can be collected from surveys which ask the respondents' information, including their age, gender, race, the type of medical treatment and the expenditure amount of each type of medical supplies that are used. In recommender systems~\cite{resnick1997recommender, bobadilla2013recommender, ricci2011introduction} in online-shopping websites which are built for product recommendations, the data can include the purchase history of the clients or the property of the products, both containing numerical and categorical features. In this paper, we focus on the problem of how to slightly perturb the input (mixed-type) data sample of a model to mislead the model's prediction. This study can have great significance to help us understand the potential risk of ML models under malicious perturbations in the applications mentioned above.

\section{Methodology - \model}\label{sec:method}

In this section, we introduce the details of our proposed method \model. Before discussing the details of \model, we first provide the necessary definitions and notations used in this work. For any given sample $x$ with true label $y\in \mathcal{Y}$, we assume that it contains the numerical features $x_n\in\mathbb{R}^{d_n}$ with dimensional $d_n$. It also contains $d_c$ categorical features $x_c$ (such as race, gender, education, etc.) and each categorical feature may include a different number of categories. More specifically, among all categorical features, we denote each individual categorical feature $x_c^{i}$ (e.g., gender), which can be chosen from $k_i$ different categories (e.g., male and female). During the attacking process, we assume that there is a classifier $F(\cdot)$ that takes the tuple $(x_n, x_c)$ as input, and makes the prediction by $F(x_n, x_c) = \hat{y} \in \mathcal{Y}$.

\subsection{An Overview of \model}

In this subsection, we introduce the overall framework of our proposed method - \model. In this paper, we only consider the scenario that all information about the model $F(\cdot)$, including its parameters and architecture, is given to the attacker. Thus, during the test phase, for a test sample $x$, which contains the numerical feature $x_n$ and categorical feature $x_c$, with true label $y$, the attacker aims to find such a perturbed data sample $x' = (x'_n, x'_c)$, which is close to $(x_n, x_c)$, but can mislead the model to make a prediction different from $y$. To achieve this goal, in our proposed \model, we aim to solve the following problem to find $(x'_n, x'_c)$ as:
\begin{align}\label{eq:overall}
\begin{split}
\max~~~~~~ & \mathcal{L}(F(x'_n, x'_c), y)\\
\text{s.t. } &
\begin{cases}
||x'_n-x_n||_1 \leq \epsilon_1\\
||x'_c-x_c||_0 \leq \epsilon_2\\
\mathcal{D}((x_n, x_c), (x'_n, x'_c))\leq \epsilon_3
\end{cases}
\end{split}
\end{align}
Specifically, we aim to find such a tuple $(x'_n, x'_c)$ where the model $F(\cdot)$ has the maximized loss value so that the model is likely to have a wrong prediction. Moreover, we introduce three constraints to regularize the generated attacks (or perturbations) to be close to the original sample $(x_n, x_c)$: 
\begin{enumerate}[(i)]
    \item We constrain the $l_1$ norm distance between the perturbed sample's numerical feature $x'_n$ and the original numerical feature $x_n$ so that the perturbation on $x_n$ is sparse. 
    \item We constrain the $l_0$ difference between $x_c$ and $x_c'$ so that the number of changed categorical features is limited. 
    \item We constrain the distance $\mathcal{D}(\cdot)$ of the pair $(x'_n, x'_c)$ to $(x_n, x_c)$ to keep the perturbed sample ``seemingly'' benign. This will help to keep the generated adversarial examples close to the distribution of clean samples. Thus, the generated samples are harder to be detected by Out-of-Distribution methods or human experts. In Section~\ref{sec:M_dist}, we will make a detailed definition about the distance $\mathcal{D}(\cdot)$.
\end{enumerate}
It is worth mentioning that: if we don't consider the constraint (iii), there already exists well-established methods to partially solve the problem in Eq.(\ref{eq:overall}). For example,  applying $l_1$-PGD method~\cite{maini2020adversarial} can solve the problem under constraint (i); and applying greedy search~\cite{yang2020greedy} can solve the problem under constraint (ii). Thus, a potential solution for Eq.(\ref{eq:overall}) is that: we first apply $l_1$-PGD~\cite{maini2020adversarial} to find $x'_n$ under constraint (i) to maximize the loss value, and then adopt greedy method~\cite{yang2020greedy} to find $x'_c$ (or vice versa). However, our proposed method, \model, is a distinct approach, which simultaneously updates the numerical and categorical features. This will help \model find better solutions for the optimization problem because it takes the interaction between numerical and categorical features into consideration. Next, we will detail how to solve Eq.(\ref{eq:overall}) -- {\it In the remaining subsections of Section~\ref{sec:method}, we will introduce how to achieve the optimization goal in Eq.(\ref{eq:overall}) which satisfies the first two constraints while ignoring the third constraint; while in Section~\ref{sec:M_dist}, we will define \distance $\mathcal{D}(\cdot)$ in detail and involve the constraint (iii) in the algorithm. }

\subsection{Sparse Adversarial Perturbations}

Overall, in our proposed \model, in order to find sparse adversarial perturbations, we introduce a new searching space, $\mathcal{S}_n\times \mathcal{S}_c$, which is the Cartesian product of the \textbf{\textit{Numerical Adversarial Space} $S_n$} where the adversarial numerical features $x'_n$ reside, and \textbf{\textit{Probabilistic Adversarial Space} $S_c$} where the adversarial categorical features $x'_c$ are sampled from.

\subsubsection{\textbf{Numerical Adversarial Space}} 
We first define the \textit{Numerical Adversarial Space} to be equal to the space of samples that satisfy the constraint (i) in Eq.(\ref{eq:overall}):
$$\mathcal{S}_n  = \{x_n' : ||x'_n - x_n||_1\leq \epsilon_1\}$$
During the \model, we repeatedly update the numerical features of $x'_n$ in the space $\mathcal{S}_n$, via an approach similar to the method Projected Gradient Descent (PGD)~\cite{madry2017towards,tramer2019adversarial} which has been designed for $l_1$ adversarial attack. 
In each step of the PGD algorithm, suppose that the objective is simply to maximize the value of a function $Q(\cdot)$, and the current optimized result is $x'_n$,
it first updates $x'_n$ along with the direction of gradient ascent:
\begin{equation}\label{eq:l1_pgd}
    x''_n \leftarrow x'_n + \text{argmax}_{||v||_1\leq \alpha}~~ v^T \nabla Q(x'_n).
\end{equation}
In other words, for each step, we find the $x''_n$ which can result in the maximal increase of the loss value. Then, if $x''_n$ is out of $\mathcal{S}_n$, we project $x''_n$ back to the defined numerical adversarial space $\mathcal{S}_n$ by solving a convex optimization problem:
$$x_n' \leftarrow \text{argmin} ||x_n'' - x_n'||_2, ~~\text{s.t.}~~ x_n' \in ||x_n' - x_n||\leq \epsilon. $$ 
It finds the point residing in $\mathcal{S}_n$ which has the smallest Euclidean distance with $x''_n$.
To solve this convex optimization problem, we can apply existing convex optimization solvers from tools like CVXOPT~\cite{diamond2016cvxpy} or the method~\cite{duchi2008efficient}.

\begin{figure*}[t]
    \centering
    \includegraphics[width=0.6\textwidth]{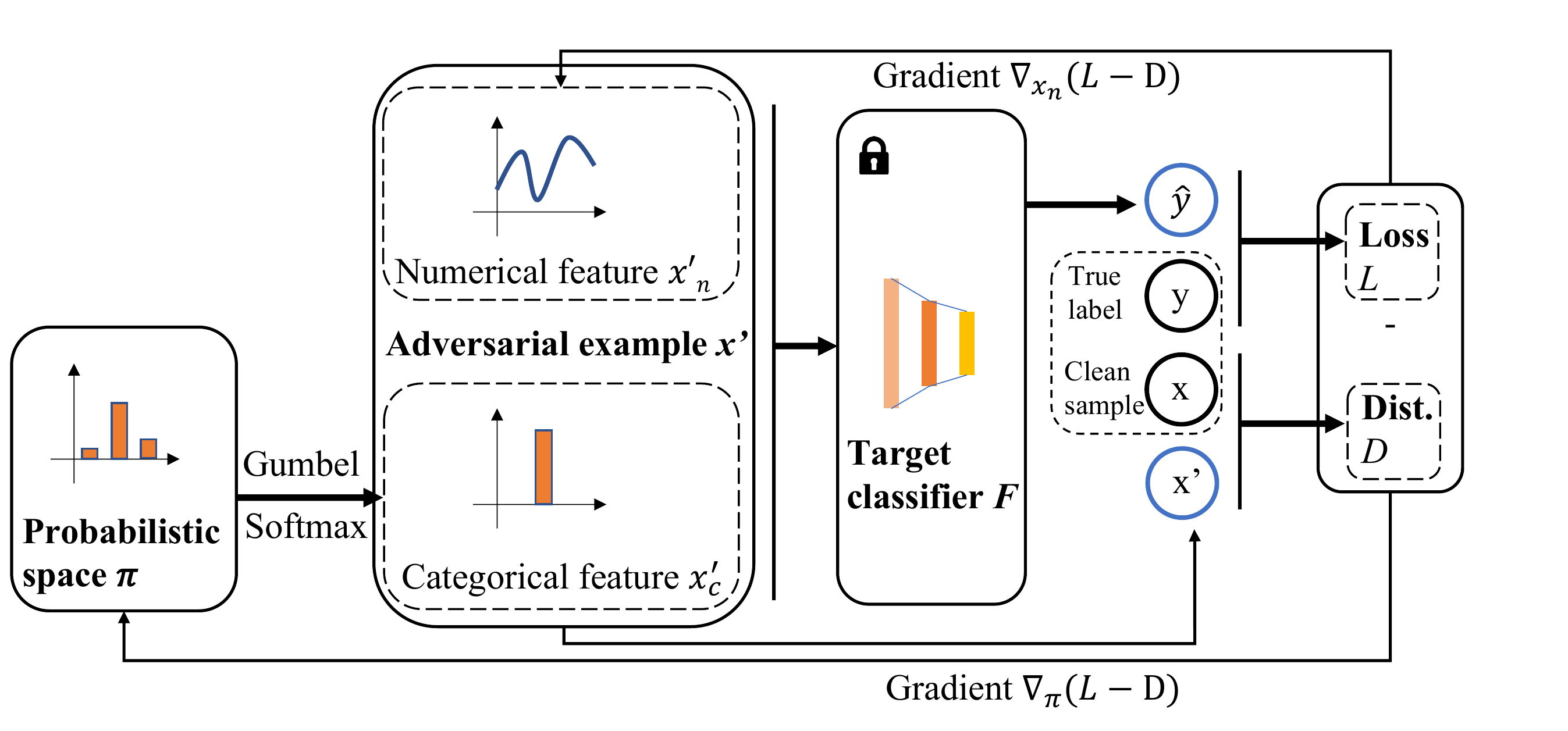}
    \caption{An overview of the proposed \model. In the forward propagation, \model generates categorical adversarial perturbation $x'_c$ from Probabilistic Distribution $\pi$, and feeds the mixed-type adversarial example $(x'_n, x'_c)$ into the targeted classifier. In the backward propagation, \model updates $\pi$ and numerical feature $x'_n$ simultaneously.}
    \label{fig:frame}
\end{figure*}

\subsubsection{\textbf{Probabilistic Adversarial Space (For Categorical Feature)}} 
To find adversarial perturbations $x'_c$ for categorical features, we propose a method to alternatively search in a continuous space $\mathcal{S}_c$, so that we can apply gradient methods to update $x'_n$ and $x'_c$ at the same time.
In detail, we assume that each feature of (adversarial) categorical data $x_c^i$ follows a distinct categorical distribution: $\textit{Categorical}(\pi_i)$, where: $$\pi_i = [\pi_{i,0}, \pi_{i,1}, ..., \pi_{i, k_i}] \in (0,1)^{k_i} = \Pi_i,$$  
$k_i$ is the number of categories of feature $x_c^i$. Here, each $\pi_{i,j}$ represents the probability that the feature $x_c^i$ belongs to the category $j$.
Thus, the distribution of all categorical features in one sample can be represented as an array $\pi = [\pi_0; \pi_1; ...;\pi_{d_c}] \in \Pi_0 \times \Pi_1 \times ...\times \Pi_{d_c} = \Pi$ , and $d_c$ is the number of categorical features.
Then, in order to constrain the $l_0$ distance between $x'_c$ and $x_c$ in constraint (ii) of Eq.(\ref{eq:overall}),
we define the Probabilistic Adversarial Space $\mathcal{S}_c$ as:
\begin{equation}\label{prob cons}
\mathcal{S}_c =  \left\{ \pi:  \Pr_{x'_c\sim\pi}(\|x'_c-x_c\|_0 \geq \epsilon_2) \le \delta \right\} \subseteq \Pi
\end{equation}
where $\delta$ is the tail probability constraint. Intuitively, by constraining $\pi$ in space $\mathcal{S}_c$, we aim to find a distribution $\pi$, 
such that for a sample $x_c'$ following the distribution $\pi$, it has a low probability to have the $l_0$ distance $||x'_c - x_c||_0$ larger than $\epsilon_2$. It is approximately equivalent to  the constraint (ii) in Eq.(\ref{eq:overall}).
In our algorithm of \model, we substitute the $l_0$ distance between $x'$ and $x$ by calculating the sum of Cross Entropy Loss between $\pi_i$ and $x_c^i$: $\mathcal{L}_{CE}(\pi_i, x_c^i)$ for all features $i\in \{1,...,d_c\}$. 
\begin{equation}\label{eq:step1}
   \Pr_{x'_c\sim\pi}(||x'_c-x_c||_0\geq\epsilon_2) \rightarrow \left[\sum_{i} \mathcal{L}_{CE}(\pi_i, x_c^i) -\zeta\right]^+ := \mathcal{L}_{CE}(\pi, x_c)
\end{equation}
It is because  $\mathcal{L}_{CE}(x_c^i, \pi_i)$ approximately measures the possibility that the generated $(x'_c)^i$ (following $\pi$) is different from $x_c^i$. Therefore, we use the sum of Cross Entropy $\sum_{i} \mathcal{L}_{CE}(\pi_i, x_c^i)$ to approximate the number of changed features, which is the $l_0$ difference $||x_c'-x_c||_0$. In our algorithm, we penalize the searched $\pi$ when the term $\sum_{i} \mathcal{L}_{CE}(\pi_i, x_c^i)$ exceeds a positive value $\zeta$. In this case, we equivalently limit the probability that the generated samples $x_c'$ have the number of perturbed features larger than $\epsilon_2$. During the optimization process, we can update the distribution $\pi$ via gradient-based methods such as Gumbel-softmax~\cite{DBLP:conf/iclr/JangGP17}.

\subsection{The Attack Framework}
With previous components of \model, we eventually transform the original problem in Eq.(\ref{eq:overall}) into a problem of finding a pair $(x_n', \pi)$ residing in the space $\mathcal{S}_n\times \mathcal{S}_c$ satisfying:
\begin{align}\label{eq:obj1}
\begin{split}
\max~~~~~~ & \mathbb{E}_{x'_c\sim \pi} \left[\mathcal{L}(F(x'_n, x'_c), y)\right]\\
\text{s.t. } 
& x'_c \in \mathcal{S}_n,~~~~ \pi \in \mathcal{S}_c
\end{split}
\end{align}
The term $\pi$ represents the categorical distribution of (the categorical part of) adversarial examples. Thus, in Eq.(\ref{eq:obj1}), we instead maximize the expected loss value of samples $x'_c$ which are generated following $\pi$. Moreover, for the constraint $\pi\in\mathcal{S}_c$, the Cross-Entropy Loss $\mathcal{L}_{CE}(\pi, x_c)$ used in Eq.(\ref{eq:step1}) is differentiable in terms of $\pi$. 
Thus, we further rewrite the problem to its Lagrangian form:
\begin{align}\label{eq:q}
\begin{split}
\max~~~~~~ & \mathbb{E}_{x'_c\sim \pi} \left[\mathcal{L}(F(x'_n, x'_c), y)- \alpha \mathcal{L}_{CE}(\pi, x_c)\right] = Q(x'_n, \pi)\\
\text{s.t. } & x'_c \in \mathcal{S}_n,~~~~ \pi \in \mathcal{S}_c,
\end{split}
\end{align}
where we denote the objective as $Q(x'_n, \pi)$. In the algorithm of \model, we apply a gradient descent method to repeatedly update the pair $(x'_n, \pi)$ to solve the problem above.
In detail, refer to the Algorithm \ref{algorithm} (or Figure~\ref{fig:frame}), in Step 1 and Step 2, we first calculate the gradients of the loss towards the input sample $x_n$ and the latent probability $\pi$ (for categorical features). In Step 3 and Step 4, we project the updated features back to the constrained space. During the test phase, we make a series of sampling and choose the optimal sampling which has the maximal $Q(\cdot)$.
It is worth mentioning that our method simultaneously updates the numerical and categorical features and the attacking algorithm can enjoy better effectiveness and efficiency than existing methods.  


\begin{algorithm}
\caption{Adversarial Attacks for Mixed-Data \model}\label{algorithm}
\begin{algorithmic}
\STATE  \textbf{Input:} Objective function $Q(\cdot)$ defined in Eq.(\ref{eq:q}), a sample $x$, true label $y$, number of steps of gradient descent $N$.
\STATE Initialize $x'_n = x_n, \pi = x_c$
\FOR{t = $1$ to $N$}
\STATE 1. Update $x'_n$ by gradient descent via Eq.(\ref{eq:l1_pgd})
\STATE 2. Update $\pi$ by gradient descent: $\pi \leftarrow \pi + \gamma \nabla_{\pi} Q(x'_n, \pi)  .$
\STATE 3. Project $x'_n$ back to $\mathcal{S}_n$.
\STATE 4. Project $\pi$ back to the space $\Pi$. 
\ENDFOR
\STATE  \textbf{Output:} The tuple $(x'_n, \pi)$. The eligible categorical features are sampled following the probability distribution $\pi$.
\end{algorithmic}
\end{algorithm}

\section{Mixed Mahalanobis Distance}\label{sec:M_dist}

\begin{figure}[t]
    \centering
    \includegraphics[width= 0.9\linewidth]{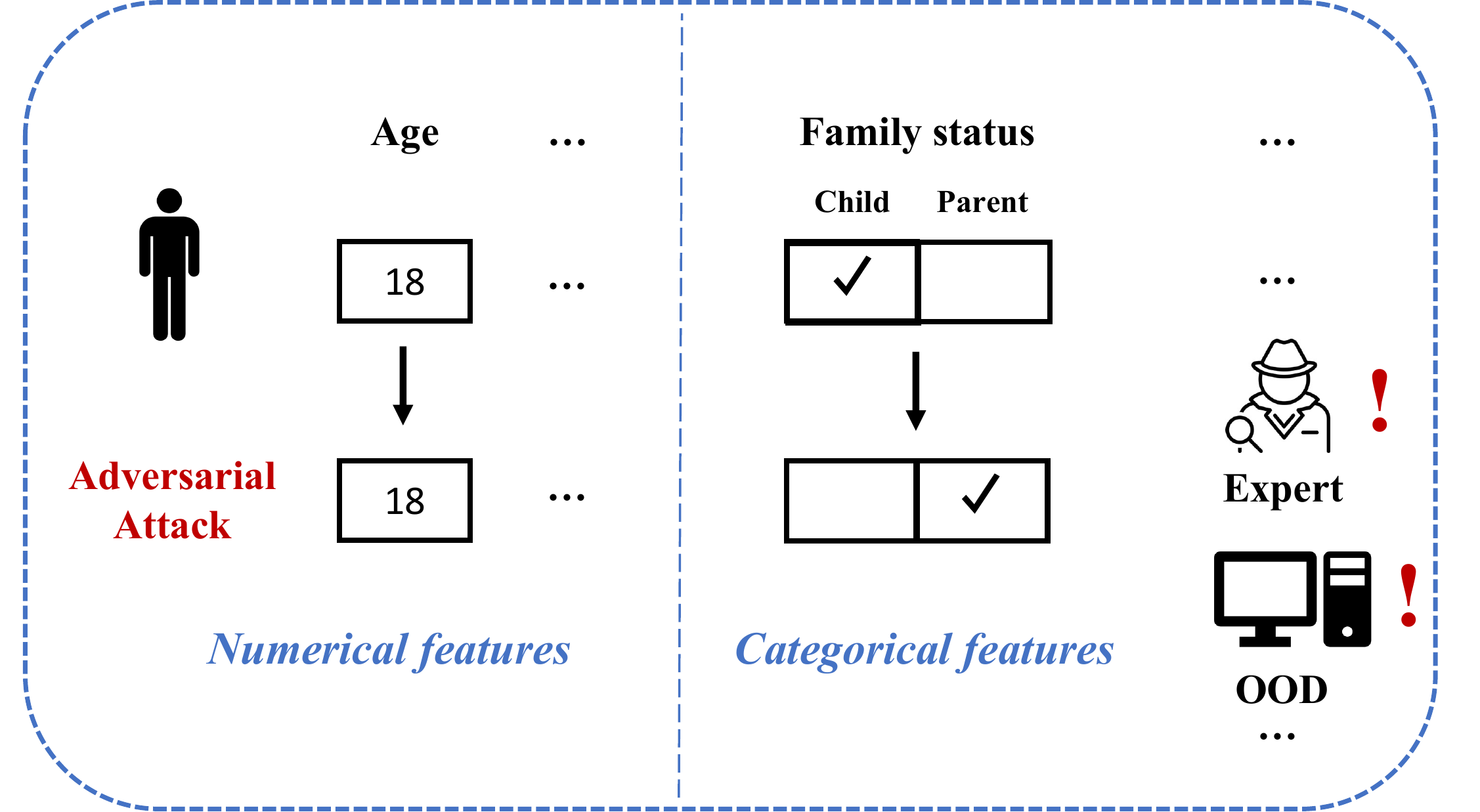}
    \caption{Out-of-Distribution Adversarial Examples can be Easily Detected, by human experts or OOD detection.}
    \label{fig:my_label}
\end{figure}

Yet solving the problem in Eq.(\ref{eq:obj1}) to satisfy the constraints (i) and (ii) helps us find sparse perturbations, it is still not adequate to guarantee the adversarial perturbations unnoticeable. For example, take Figure~\ref{fig:my_label} as an example, an adversarial attacker perturbs the feature ``family status'' from ``child'' to ``parent'' for an 18-year-old loan applicant (in Home Credit Dataset). The perturbed sample can significantly deviate from the clean data distribution because there are rarely 18-year-old parents in reality. As a result, the adversarial perturbation can be easily detected as ``abnormal samples'' by human experts and they can refuse to let ML models predict these ``abnormal'' samples. Moreover, these abnormal samples are also very likely to be detected by Out-of-Distribution (OOD) detection methods~\cite{botev2010kernel, ren2019generating}. For example,  the methods~\cite{botev2010kernel, ren2019generating} first estimate the distribution of clean data samples via kernel density estimation or via training generative models and calculate the likelihood of a test sample in the estimated clean data distribution. Similarly, if a sample's likelihood score is low, the model can refuse to make a prediction. It is worth mentioning that OOD detection is widely applied as defense strategies~\cite{uwimana2021out} in image classification tasks. Therefore, during the attack process, the attacker should also consider generating adversarial examples, which are close to the clean data distribution. In other words, we also need to consider the constraint (iii) in Eq.(\ref{eq:obj1}).

\subsection{Mahalanobis Distance}

In order to generate in-distribution adversarial examples, the method in
\cite{erdemir2021adversarial} considers regularizing the Mahalanobis Distance~\cite{de2000mahalanobis, mclachlan1999mahalanobis} of the generated adversarial examples to the clean distribution (for numerical data). In detail, suppose that the clean data samples follow a Gaussian distribution $\mathcal{P} = \mathcal{N}(\mu, \Sigma)$ with the center $\mu$ and covariance matrix $\Sigma$, the Mahalanobis Distance measures a sample $x$'s distance to this distribution $\mathcal{P}$ by calculating:
\begin{equation}\label{eq:maha}
    \mathcal{D}(x, \mathcal{P}) = (x - \mu)^T \Sigma^{-1} (x-\mu)
\end{equation}
Intuitively, the smaller Mahalanobis Distance the sample $x$ has, the larger density score it will receive. Note that the density function of the distribution $\mathcal{P}$ is:
$
p(x, \mathcal{P}) \propto \exp(-\frac{1}{2}(x-\mu)^T\Sigma(x-\mu)).$
Therefore, a sample $x$ with a higher density score is more likely to belong to the distribution $\mathcal{P}$. The attack method in~\cite{erdemir2021adversarial} constrains the sample-wise Mahalanobis distance between the generated adversarial example $x'$ and the clean sample $x$, which is defined as:
\begin{equation}\label{eq:maha2}
    \mathcal{D}(x', x, \mathcal{P}) = (x' - x)^T \Sigma^{-1} (x'- x)
\end{equation}
Notably, if $x$ is a clean sample, it will be close to the clean distribution $\mathcal{P}$, so it will have a small Mahalanobis Distance $\mathcal{D}(x, \mathcal{P})$. Thus, penalizing the term $\mathcal{D}(x', x, \mathcal{P})$ will help the adversarial example $x'$ also have a small $ \mathcal{D}(x', \mathcal{P})$ (by Triangle inequality). Therefore, $x'$ is also highly likely to follow the clean distribution $\mathcal{P}$.

\begin{figure*}[t]
     \centering
\begin{adjustbox}{minipage=\linewidth,scale=0.9}
     \begin{subfigure}[b]{0.32\textwidth}
         \centering
    \includegraphics[width=\textwidth]{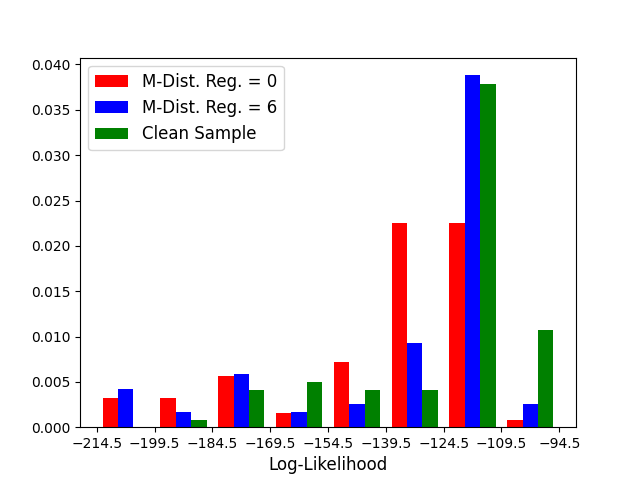}
         \caption{CIS Fraud.}
     \end{subfigure}
     \hfill
     \begin{subfigure}[b]{0.32\textwidth}
         \centering
         \includegraphics[width=\textwidth]{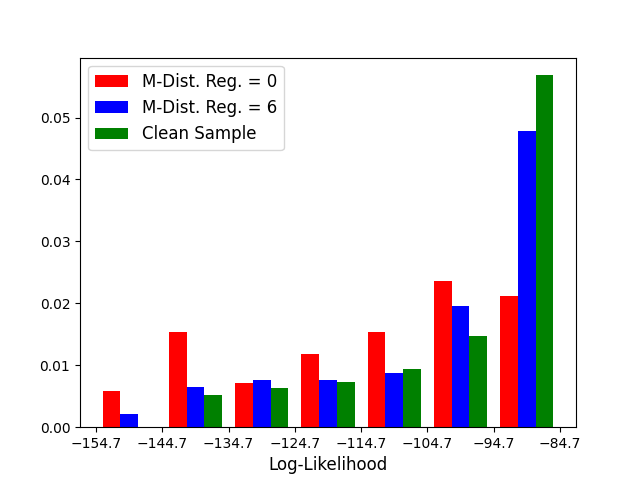}
         \caption{Home Credit.}
     \end{subfigure}
          \hfill
     \begin{subfigure}[b]{0.32\textwidth}
         \centering
         \includegraphics[width=\textwidth]{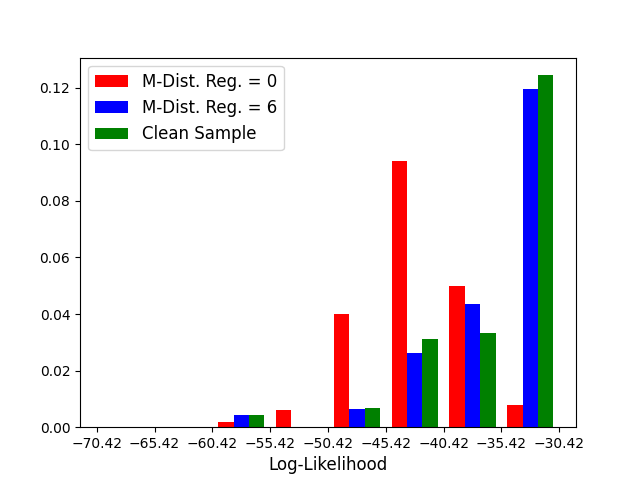}
         \caption{Criteo.}
     \end{subfigure}
        \caption{Imposing Small Mix Mahalanobis Distance Helps Generating In-Distribution Samples}
        \label{fig:exp3}
                    \end{adjustbox}
\end{figure*}

\subsection{Mixed Mahalanobis Distance}
In this work, we extend the idea of Mahalanobis Distance to mixed-type data. Overall, we first define a \textit{Generalized Covariance Matrix} specifically for mixed-type data, and then we define \distance based on the Generalized Covariance Matrix (similar to the definition in Eq.(\ref{eq:maha2})).
Specifically, we consider that the attacker has the knowledge of a sufficient amount of clean data samples and we define the clean data matrix, $X = \{(x_n, x_c)_j\}_{j = 1,...,N}$, which is composed of $N$ different clean samples. We also denote $X = (X_n, X_c)$, where each column of $X_n$ represents a numerical feature and $X_c$ denotes the categorical features.
In $X_c$, we encode each categorical feature of each sample into its one-hot representation. For example, a column in the matrix $X_c$ represents whether the (categorical) feature $i$ of these samples belongs to the category $j$ (e.g. whether the gender is 'female'). Based on this definition of $X$, we can define the \textbf{Generalized Covariance Matrix} as:
\begin{equation}
    \hat{\Sigma}_M = \frac{1}{N-1}~ \tilde{X}^T \cdot \tilde{X}
\end{equation}
where $\tilde{X}$ is the centered matrix of $X$: $\tilde{X} = X - \bar{X}$. In this Generalized Covariance Matrix $\hat{\Sigma}_M$, each element $(\hat{\Sigma}_M)_{i,j}$ measures the relation between a pair of features $(i,j)$:
\begin{enumerate}
    \item If feature $i$ is numerical and feature $j$ is numerical, it estimates the covariance between features $i$ and $j$ as 
    $$(\hat{\Sigma}_M)_{i,j} = \text{cov}(X[i], X[j])$$
     \item If feature $i$ is numerical and feature $j$ is categorical, since the column $X[j]$ is the one-hot representation which only contains $0$ and $1$, the term $(\hat{\Sigma}_M)_{i,j}$ becomes:
     $$(\hat{{\Sigma_M}})_{i,j} = \frac{N_0 N_1}{N(N-1)}(\bar{X_0}[i] - \bar{X_1}[i])$$
     where $N_0$ and $N_1$ denote the number of samples in $X$ whose feature $j$ belongs to the category $0$ and $1$ respectively; and $\bar{X_0}[i]$ and $\bar{X_1}[i]$ denote the average value of feature $i$ of samples whose feature $j$ belongs to the category $0$ and category $1$, respectively. Intuitively, this term measures the interaction between a numerical feature $i$ and a categorical feature $j$. For example, if $(\hat{{\Sigma_M}})_{i,j}$ is much larger than 0, the average value of feature $i$ of samples in category $0$ is much larger than that in category $1$. It suggests that there is a great discrepancy of feature $i$ between different categories of feature $j$.
     \item  If feature $i$ is categorical and feature $j$ is categorical, it becomes:
     $$(\hat{{\Sigma_M}})_{i,j} = \frac{N_{1,0}\cdot N_{0,1} - N_{0,0}\cdot N_{1,1}}{N\cdot (N-1)}$$
     where we use $N_{(\cdot,\cdot)}$ to denote the number of samples that belong to each individual category (for both feature $i$ \& $j$). For example, $N_{1,0}$ is the number of samples whose feature $i$ is $1$ and feature $j$ is $0$. The square of this term is widely used as a chi-squared statistic in \textit{correspondence analysis}~\cite{greenacre2017correspondence} to test the (in)-dependence of two categorical variables. 
\end{enumerate}

\noindent \textbf{Mixed Mahalanobis Distance}: 
Equipped with the Generalized Covariance Matrix, we follow the strategy similar to Eq.(\ref{eq:maha2}) to define the Mahlanobis Distance for mixed-type data. Since the clean data matrix $X$ contains one-hot representations, $X$ is not full-rank and $\hat{\Sigma}_M$ is not invertible.
Thus, we approximate the inverse covariance matrix $(\hat{\Sigma}_M)^{-1}$ by only preserving its first $m$ eigenvectors. In detail, we first conduct eigenvalue decomposition for $\hat{\Sigma}_M$:
$(\hat{\Sigma}_M)' =  V \cdot S \cdot V^{-1}$, where $V$ is the eigenvectors and $S$ is the diagonal matrix whose non-zero elements are the top $m$ eigenvalues of $\hat{\Sigma}_M$. In this way, we get the inverse covariance matrix $(\hat{\Sigma}_M)^{-1} = V\cdot S^{-1}\cdot V^{-1}$, and we have the \distance between the adversarial sample $x'$ and clean sample $x$ as:
\begin{equation}
    \mathcal{D}(x', x) = (x'-x)^T (\hat{\Sigma}_M)^{-1} (x'-x)
\end{equation}
During the algorithm of \textbf{M-Attack}, when we solve the problem in Eq.(\ref{eq:overall}), we can satisfy the constraint (iii): $\mathcal{D}(x', x)\leq \epsilon_3$ by penalizing the large $\mathcal{D}(x', x)$. In detail, because the term $\mathcal{D}(x',x)$ is differentiable in terms of $x'$, we have an alternative objective for Eq.(\ref{eq:overall}), which is defined as: 
\begin{align}\label{eq:overall2}
\begin{split}
\max~~~~~~ & \mathcal{L}(F(x'_n, x'_c), y) -\lambda\cdot \mathcal{D}((x'_n, x'_c), (x_n, x_c))\\
\text{s.t. } &
\begin{cases}
||x'_n-x_n||_1 \leq \epsilon_1\\
||x'_c-x_c||_0 \leq \epsilon_2\\
\end{cases}
\end{split}
\end{align}
\noindent where $\lambda > 0$ controls the trade-off between the loss and \distance. Facilitated with the introduced method in Section~\ref{sec:method}, we can successfully solve this problem in Eq.(\ref{eq:overall2}).

\begin{figure*}[t]
     \centering
         \begin{adjustbox}{minipage=\linewidth,scale=0.88}
     \begin{subfigure}[b]{0.32\textwidth}
         \centering
    \includegraphics[width=\textwidth]{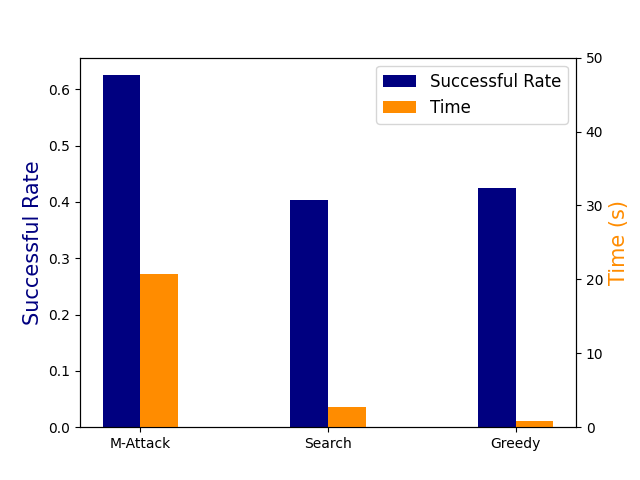}
         \caption{CIS Fraud. $(\epsilon_1 = 0.6, \epsilon_2 = 3, \lambda = 6$)}
     \end{subfigure}
     \hfill
     \begin{subfigure}[b]{0.32\textwidth}
         \centering
         \includegraphics[width=\textwidth]{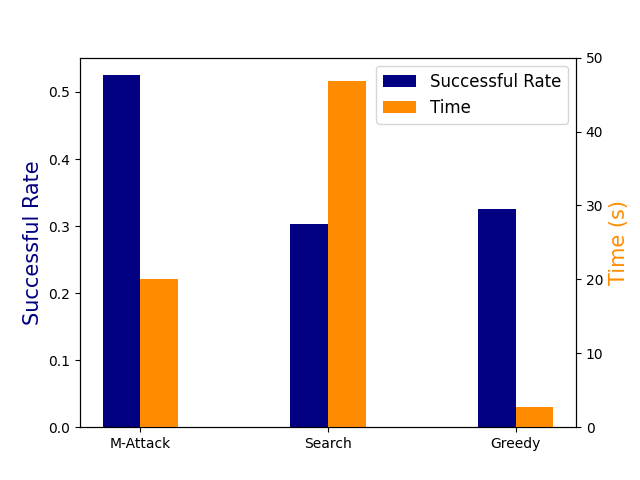}
         \caption{CIS Fraud. $(\epsilon_1 = 0.6, \epsilon_2 = 4, \lambda = 6)$}
     \end{subfigure}
     \hfill
     \begin{subfigure}[b]{0.32\textwidth}
         \centering
         \includegraphics[width=\textwidth]{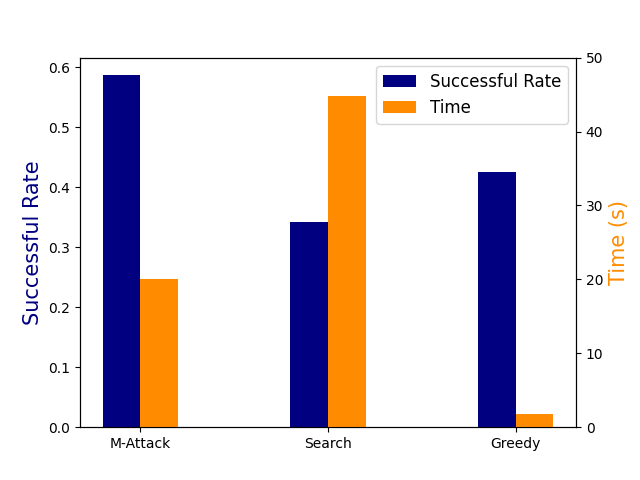}
         \caption{CIS Fraud $(\epsilon_1 = 0.6, \epsilon_2 = 4, \lambda = 10)$}
     \end{subfigure}
     \centering
     \begin{subfigure}[b]{0.32\textwidth}
         \centering
    \includegraphics[width=\textwidth]{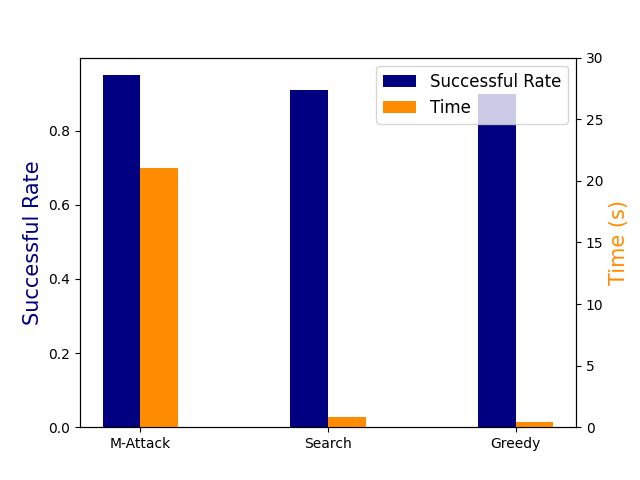}
         \caption{Home Credit. $(\epsilon_1 = 0.3, \epsilon_2 = 2, \lambda = 6)$}
     \end{subfigure}
     \hfill
     \begin{subfigure}[b]{0.32\textwidth}
         \centering
         \includegraphics[width=\textwidth]{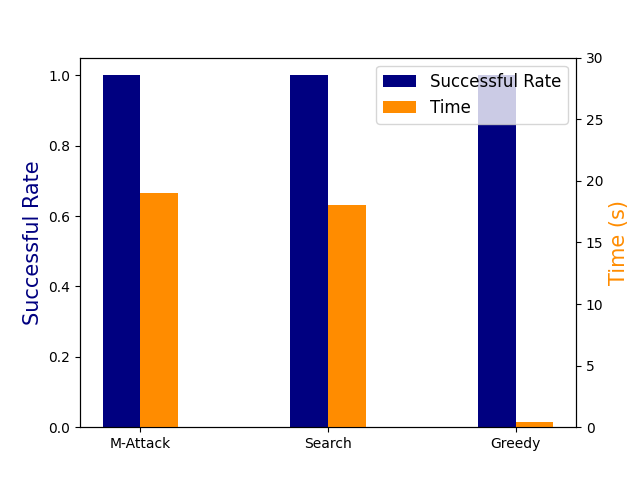}
         \caption{Home Credit. $(\epsilon_1 = 0.3, \epsilon_2 = 4, \lambda = 6)$}
     \end{subfigure}
     \hfill
     \begin{subfigure}[b]{0.32\textwidth}
         \centering
         \includegraphics[width=\textwidth]{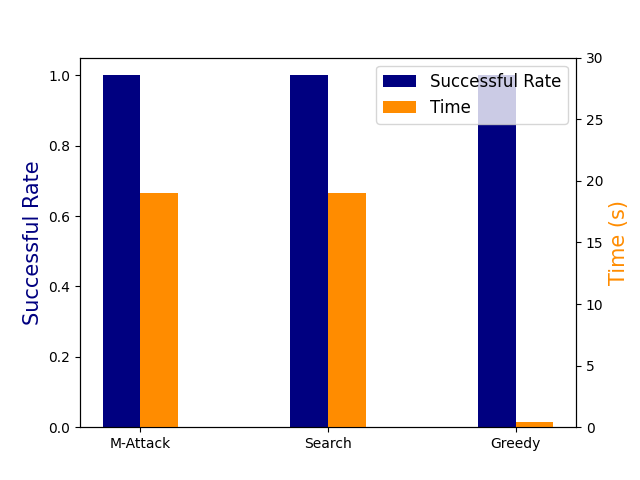}
         \caption{Home Credit. $(\epsilon_1 = 0.3, \epsilon_2 = 4, \lambda = 10)$}
     \end{subfigure}
          \begin{subfigure}[b]{0.32\textwidth}
         \centering
    \includegraphics[width=\textwidth]{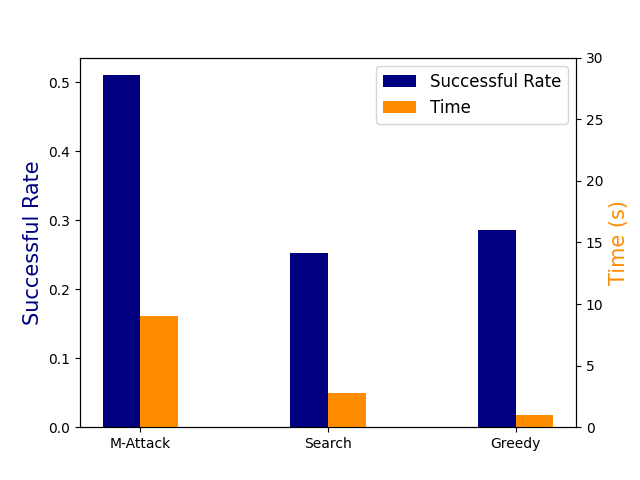}
         \caption{Criteo. $(\epsilon_1 = 0.6, \epsilon_2 = 3, \lambda = 2)$}
     \end{subfigure}
     \hfill
     \begin{subfigure}[b]{0.32\textwidth}
         \centering
         \includegraphics[width=\textwidth]{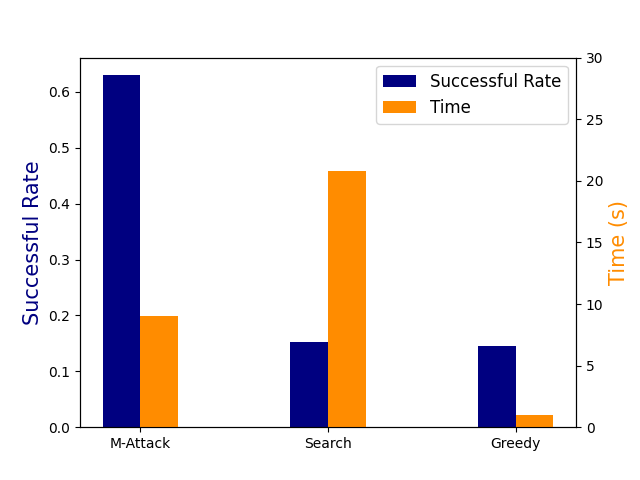}
         \caption{Criteo. $(\epsilon_1 = 0.6, \epsilon_2 = 4, \lambda = 2)$}
     \end{subfigure}
     \hfill
     \begin{subfigure}[b]{0.32\textwidth}
         \centering
         \includegraphics[width=\textwidth]{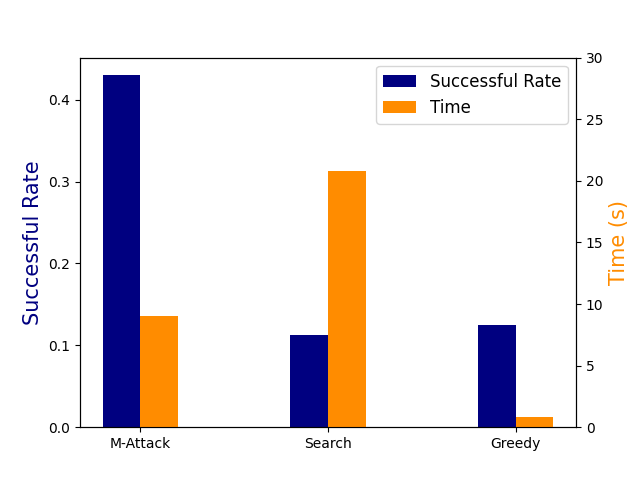}
         \caption{Criteo. $(\epsilon_1 = 0.6, \epsilon_2 = 4, \lambda = 4)$}
     \end{subfigure}
        \caption{\model vs. Baselines by comparing Computational Cost and Attack Successful Rate}
        \label{fig:exp2}
    \end{adjustbox}
\end{figure*}

\section{Experiment}\label{sec:exp}
In this section, we validate the effectiveness of our proposed \model and \distance. In detail, in Section~\ref{OOD}, we first verify that imposing a small \distance helps generating in-distribution adversarial examples. Then,
we compare the performance \model with baseline attack methods in Section~\ref{sec:efficiency} and Section~\ref{sec:optimality}. 
Before introducing the experimental results, we first introduce the basic experimental settings and baselines. 

\subsection{Experimental Settings}
In this subsection, we introduce the datasets and baseline methods studied in this paper.

\subsubsection{Datasets} We mainly conduct the experiments on three datasets, IEEE-CIS transaction dataset\footnote{\url{https://www.kaggle.com/competitions/ieee-fraud-detection}}, Home Credit Default Risk Dataset\footnote{\url{https://www.kaggle.com/competitions/home-credit-default-risk}},  and Criteo Display Advertising dataset\footnote{\url{https://www.kaggle.com/c/criteo-display-ad-challenge}}. For all these datasets, we only preserve the categorical features which have no more than $50$ categories:
\begin{itemize}
\item IEEE-CIS transaction data contains transaction records from e-commerce, which contain the transaction summary and identify information of clients. The task is to predict if a transaction is benign or fraudulent.  It contains 108 numerical features and 32 categorical features.
\item Home Credit Default Risk contains historic records of loan applicant's demographic information and bank account information. The goal is to predict whether a loaner will pay back his/her loan.  It contains 73 numerical features and 28 categorical features.
\item The Criteo Display Advertising Challenge Dataset is about solving the machine learning task: given a user and the webpage he is visiting, whether he will click on a given advertisement. It contains 13 numerical features and 7 categorical features.
\end{itemize}

For all datasets, we train a two layer Multilayer Perceptron (MLP) model with 64 hidden nodes for the classification tasks, and we will leave other popular classifiers, such as tree based methods~\cite{chen2016xgboost, belgiu2016random} for one future work.

\subsubsection{Baselines.} To the best of our knowledge, there is no existing (popular) baseline methods for the studied problem. Therefore, we compare our proposed \model with the combination of existing attack methods to sequentially perturb the categorical features and numerical features. In detail, we consider baseline methods:
\begin{itemize}
    \item $l_1$ PGD + Search: It first applies $l_1$ PGD~\cite{maini2020adversarial} to perturb numerical features. Then, it adopts the search method~\cite{bao2021towards} to perturb categorical features. The search method first perturbs each categorical one-by-one and selects the top $\epsilon_2$ features with the highest impact on the model output. Then, it traverses all combinations of these $\epsilon_2$ features to find the perturbation with the highest loss value.
    \item $l_1$ PGD + Greedy: It also conducts $l_1$ PGD attack and then finds the $\epsilon_2$ features with the highest impact on the model output. Then, it perturbs these $\epsilon_2$ features greedily. For each feature, it assigns this feature with the category which leads to maximal loss and keeps the other perturbed features fixed, so it avoids searching all combinations.
\end{itemize}
Note we can also first perturb the categorical features, and then perturb the numerical features. We find this strategy has the similar empirical performance with our considered baselines, so we only report the performance ``$l_1$ PGD + Search'' \& ``$l_1$ PGD + Greedy''.

\subsection{Mix Mahalanobis Distance Helps Generating In-Distribution Samples}\label{OOD}

In this experiment, we first verify that regularizing our proposed \distance can effectively help the generated adversarial examples to be closer to the clean data distribution. To confirm this point, in all three datasets, we follow the strategy similar to~\cite{ren2019generating, xiao2018generating} to first estimate the distribution of clean data samples, then we test whether the generated adversarial examples is clearly deviated from our estimated clean data distribution. Specifically, in our experiment in Figure~\ref{fig:exp3}, we apply the Kernel Density Estimation~\cite{botev2010kernel} to estimate the distribution of all clean training samples. Then, in Figure~\ref{fig:exp3}, we calculate the estimated Log-Likelihood (density score) of clean test samples (green), as well as the generated adversarial examples (via \model) with a strong Mix Mahalanobis Distance regularization $\lambda = 6$ (blue), and adversarial examples with no Mix Mahalanobis Distance regularization $\lambda = 0$ (red), and report their frequency in each interval of Log-Likelihood. From the result, we can see that the most clean test samples (green) have a high Log-Likelihood value, which suggests that they are close to the estimated clean data distribution. For adversarial examples which generated via strong Mahalanobis Distance regularization (blue), most samples also have high Log-Likelihood. However, for adversarial examples which generated without Mahalanobis Distance regularization (red), they have significantly lower Log-Likelihood values. Thus, they are significantly deviated from our estimated clean distribution. As a conclusion, this experimental results demonstrate that imposing a small Mix Mahalanobis Distance can help the generated adversarial examples closer to the clean distribution, which can facilitate the samples evade potential defections.

\begin{figure*}[t]
     \centering
    \begin{adjustbox}{minipage=\linewidth,scale=0.87}
     \begin{subfigure}[b]{0.32\textwidth}
         \centering
    \includegraphics[width=\textwidth]{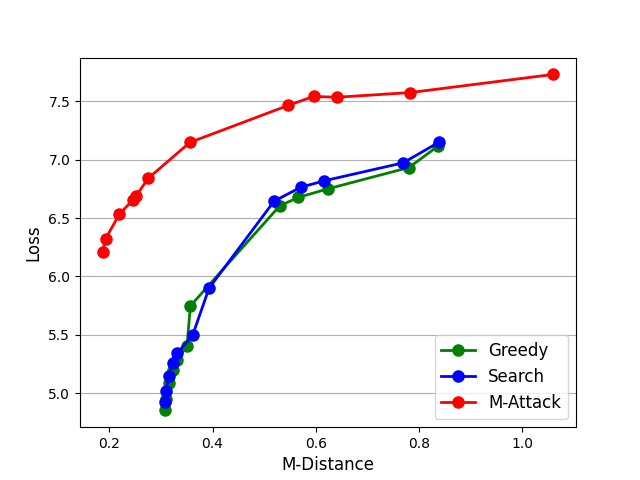}
         \caption{CIS Fraud. $(\epsilon_1 = 0.3, \epsilon_2 = 2)$}
     \end{subfigure}
     \hfill
     \begin{subfigure}[b]{0.32\textwidth}
         \centering
         \includegraphics[width=\textwidth]{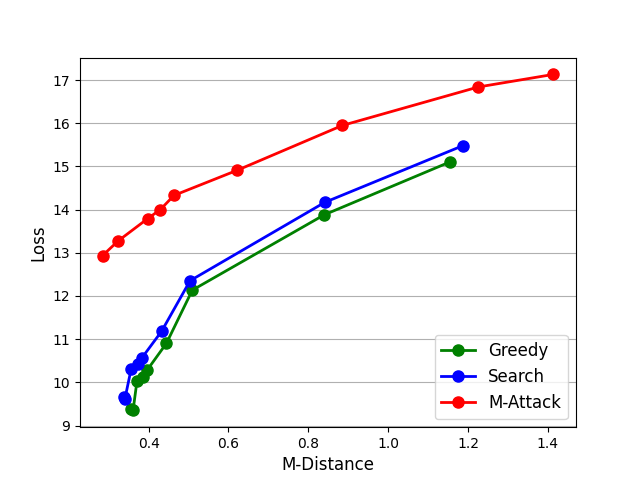}
         \caption{CIS Fraud. $(\epsilon_1 = 0.3, \epsilon_2 = 4)$}
     \end{subfigure}
     \hfill
     \begin{subfigure}[b]{0.32\textwidth}
         \centering
         \includegraphics[width=\textwidth]{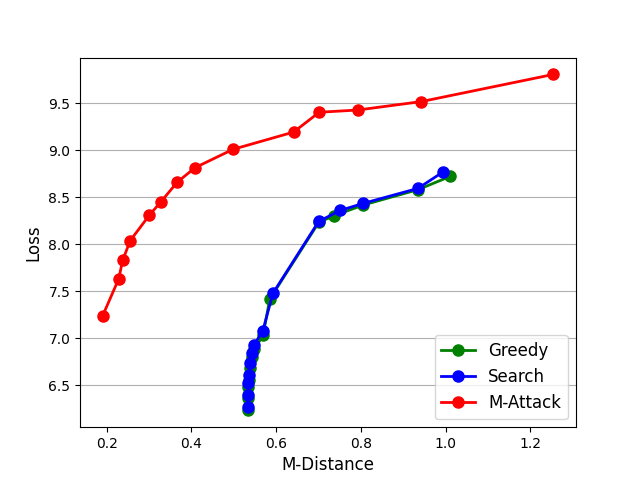}
         \caption{CIS Fraud. $(\epsilon_1 = 0.6, \epsilon_2 = 2)$}
     \end{subfigure}
     \centering
     \begin{subfigure}[b]{0.32\textwidth}
         \centering
    \includegraphics[width=\textwidth]{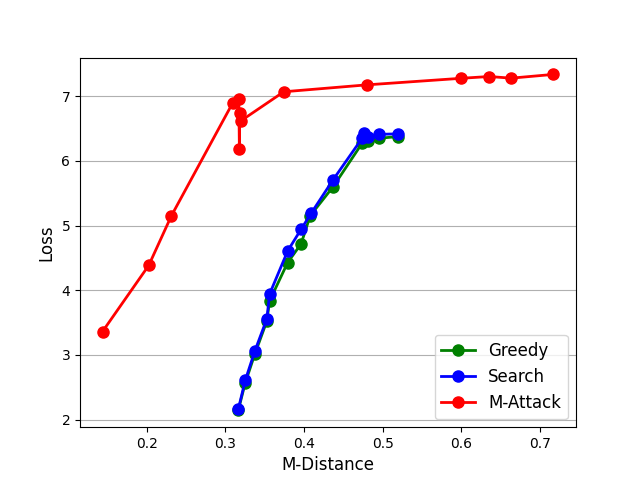}
         \caption{Home Credit. $(\epsilon_1 = 0.3, \epsilon_2 = 2)$}
     \end{subfigure}
     \hfill
     \begin{subfigure}[b]{0.32\textwidth}
         \centering
         \includegraphics[width=\textwidth]{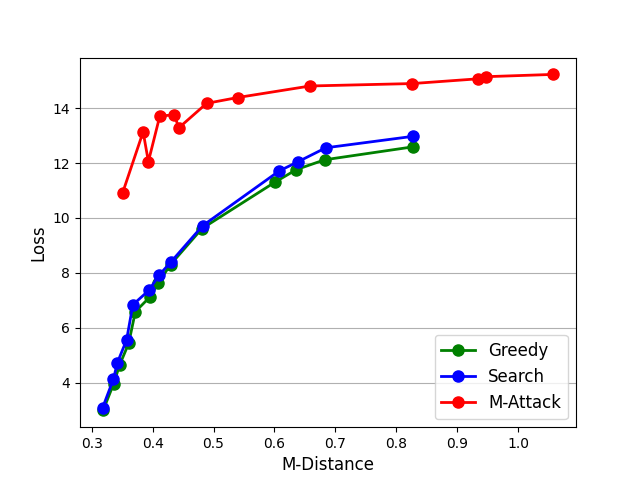}
         \caption{Home Credit. $(\epsilon_1 = 0.3, \epsilon_2 = 4)$}
     \end{subfigure}
     \hfill
     \begin{subfigure}[b]{0.32\textwidth}
         \centering
         \includegraphics[width=\textwidth]{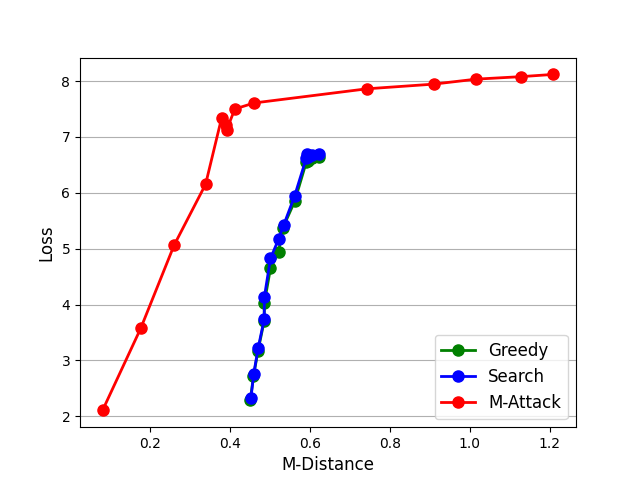}
         \caption{Home Credit. $(\epsilon_1 = 0.6, \epsilon_2 = 2)$}
     \end{subfigure}
     \centering
     \begin{subfigure}[b]{0.32\textwidth}
         \centering
    \includegraphics[width=\textwidth]{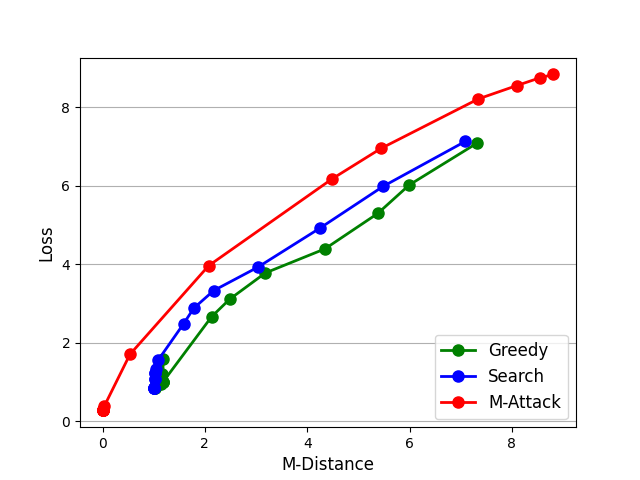}
         \caption{Criteo. $(\epsilon_1 = 0.3,\epsilon_2 = 2)$}
     \end{subfigure}
     \hfill
     \begin{subfigure}[b]{0.32\textwidth}
         \centering
         \includegraphics[width=\textwidth]{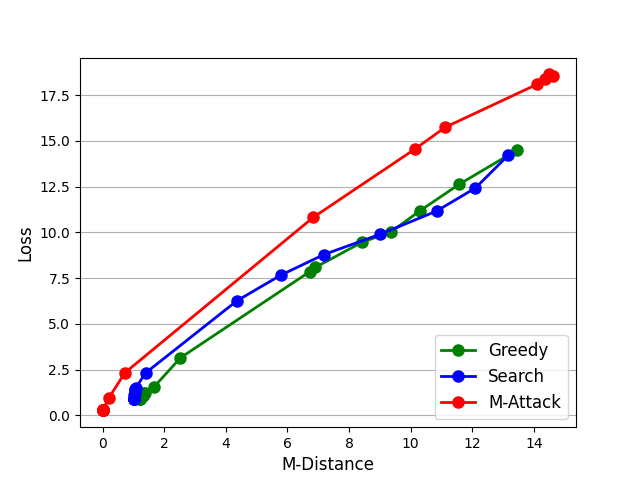}
         \caption{Criteo. $(\epsilon_1 = 0.3,\epsilon_2 = 4)$}
     \end{subfigure}
     \hfill
     \begin{subfigure}[b]{0.32\textwidth}
         \centering
         \includegraphics[width=\textwidth]{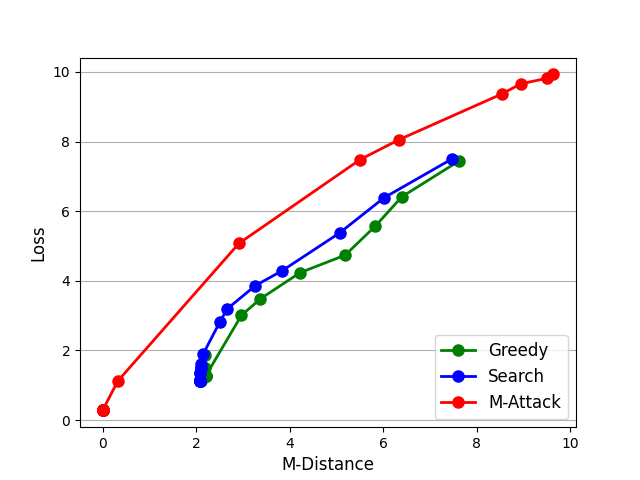}
         \caption{Criteo. $(\epsilon_1 = 0.6,\epsilon_2 = 2)$}
     \end{subfigure}
        \caption{M-Attack vs. Baselines by comparing M-Distance and Loss Trade-off}
        \label{fig:exp1}
            \end{adjustbox}
\end{figure*}

\subsection{Effectiveness and Efficiency of \model}\label{sec:efficiency}

Based on the discussion in Section~\ref{OOD}, we conduct experiments to verify whether \model can evade the OOD detection methods and successfully attack the model. In this paper, we consider the OOD detection method is based on the method introduced in Section~\ref{OOD}: if a sample has a low Log-Likelihood value (smaller than the 10-th percentile of clean samples), we flag it as ``abnormal'' and refuse prediction. In Figure~\ref{fig:exp2}, 
we report the successful attacking rate for each method: for a test sample which the classifier $x$ can correctly classify, if the model has a wrong prediction after adversarial perturbation, and it is also not flagged as ``abnormal'', we count it as a successful attack. Meanwhile, we also record the average computing time to generate each adversarial example (in seconds), where all experiments are conducted in a single NVIDIA Tesla K80 GPU.  From the results in Figure~\ref{fig:exp2}, we can see that \model has higher successful rate compared with baseline methods in CIS and Criteo dataset. The Home Credit dataset is easier to be attacked, so all methods can achieve high successful rate. Moreover, in each dataset, we usually only need perturb a few features to achieve a high successful attacking rate. In terms of computing cost, the ``$l_1$ PGD + greedy'' method spend shorter time than other methods. However, \model's computing cost will not increase with the increase of perturbation budget $\epsilon_2$, because it always conducts a fixed number of gradient steps.

\vspace{-0.5cm}
\subsection{Optimality of \model}\label{sec:optimality}

To have a deeper understanding of the behavior of \model, in the experiment in Figure~\ref{fig:exp1}, we compare \model with baselines, in terms of their optimality to solve the main problem in Eq.(\ref{eq:overall}).  In Figure~\ref{fig:exp1}, in each subfigure (with a fixed $\epsilon_1$ and $\epsilon_2$), each single dot reports the average loss value $\mathcal{L}(F(x'), y)$ of targeted classifier and the average Mix Mahalanobis Distance $\mathcal{D}(x',x)$ for 100 randomly chosen test samples, where the adversarial samples $x'$ are obtained via solving the Eq.(\ref{eq:overall2}) with a fixed value $\lambda$ (which controls the regularization of Mix Mahalanobis Distance). For baseline methods, we can also control their Mix Mahalanobis Distance by replacing their objective $\mathcal{L}(F(x'), y)$ with  $\mathcal{L}(F(x'), y) - \lambda \mathcal{D}(x',x)$. In the Figure~\ref{fig:exp1}, we report the performance with different choices of $\lambda$ for each method, to demonstrate the trade-off relation between the maximized loss value and the Mix Mahalanobis Distance. From the result, we can see when $\epsilon_1$ and $\epsilon_2$ are fixed, when different attack methods achieve the similar Mix Mahalanobis Distance, our proposed \model can always achieve the highest average loss value. Especially when the average Mix Mahalanobis Distance is lower, the advantage of \model over baselines is more obvious. This suggests that the adversarial examples generated via \model can have better chance to attack the targeted classifier and preserve the Mix Mahalanobis Distance. It is because \model simultaneously updates the numerical and categorical perturbations, so it enjoys the benefit to better optimize the term such as Mix Mahalanobis Distance. Since the categorical features and numerical features are highly interactive in Mix Mahalanobis Distance, it is sub-optimal to optimize them sequentially, as the baseline methods.

\vspace{-0.3cm}
\section{Conclusion}
\par To fill in the research gap on generating adversarial examples with mixed-type data, in this work, we propose \model, which is designed to attack both the numerical and the categorical features in a unified gradient-based algorithm. The experimental results on three real-world datasets illustrate that our proposed M-Attack can efficiently and effectively find close-to-true-distribution adversarial examples.

\balance
\bibliographystyle{unsrt}
\bibliography{reference}

\end{document}